\documentclass[letterpaper, 10 pt, conference, bookmarks=true]{ieeeconf}
\IEEEoverridecommandlockouts
\usepackage{times}

\usepackage{multicol}

\usepackage{algorithm}
\usepackage{algpseudocode}
\usepackage{amsthm,amsmath,amssymb}
\usepackage{graphicx}
\usepackage{import}
\usepackage{dsfont}
\usepackage{todonotes}
\usepackage{amsfonts}

\usepackage{mathtools}
\usepackage{acronym}
 
\usepackage{tabularx}
\let\labelindent\relax
\usepackage{enumitem}
\usepackage[most]{tcolorbox}
\usepackage{diagbox}
\usepackage[utf8]{inputenc}
\usepackage{multirow}
\usepackage{booktabs}
\usepackage{tabularx}

\DeclareMathOperator*{\argminB}{argmin}

\usepackage{svg}
\usepackage{cite}

\newcommand{\realfield}[1]{\hbox{I \kern -.5em R}^{#1}}
\newcommand {\mb}[1]{\mathbf{#1}}
\newcommand {\bs}[1]{\boldsymbol{#1}}
\newcommand{\uvec}[1]{\hat{\mathbf{#1}}}
\newcommand{\Rot}[2]{{^{#1}\mathbf{R}}_{#2}}  

\newcommand{\invT}{^{-\!\mathrm{T}}}

\title{\LARGE \bf
Sensorless Remote Center of Motion Misalignment Estimation}
\author{Hao Yang$^1$,  Lidia Al-Zogbi$^1$, Ahmet Yildiz$^2$,  Aimal Khan$^3$,
 Nabil Simaan$^2$, Jie Ying Wu$^1$%
\thanks{Authors$^1$ are with the Department of Computer Science, authors$^2$ are with the Department of Mechanical Engineering, Vanderbilt University, TN 37212, USA, and author$^3$ is with the Department of Surgery, Vanderbilt University Medical Center, TN 37232, USA. 
All correspondence should be addressed to Hao Yang {\tt\small hao.yang@vanderbilt.edu}}
}

\begin{document}

\maketitle
\thispagestyle{empty}
\pagestyle{empty}

\begin{abstract}
Laparoscopic surgery constrains instrument motion at the patient's incision point to minimize tissue trauma. Surgical robots achieve this through remote center of motion (RCM) constraints. Maintaining accurate RCM alignment with the incision site is challenging due to patient motion and tissue deformation, and such misalignment can generate unsafe forces at the incision site. This paper presents a sensorless force estimation-based framework for dynamically estimating RCM misalignment in robotic surgery. Experimental results show that misalignments greater than 20\,mm can generate forces large enough to risk tissue damage, underscoring the importance of precise RCM positioning. For misalignment $D\,\textgreater$ 20\,mm, our optimization algorithm estimates the RCM offset with an average absolute error of 4.2$\pm$2.1\,mm. Accurate RCM misalignment estimation is a step toward automated RCM misalignment compensation, enhancing safety and reducing tissue damage in robotic-assisted laparoscopic surgery.
    
\end{abstract}

\section{Introduction}
\label{sec:introduction}

Laparoscopic surgery has become the gold standard for minimally invasive abdominal and pelvic procedures, providing significant advantages over traditional open techniques. Advantages include reduced pain, recovery time, risk of incisional site hernia and infection, and financial burden  \cite{velanovich2000laparoscopic, varela2010laparoscopic,taylor2020chapter}. This approach enables surgeons to access the abdominal and pelvic cavity through small incision ports, where trocars serve as conduits for surgical tools and imaging equipment. In robotic laparoscopy, the remote center of motion (RCM) is a fundamental principle that establishes a fixed pivot constraint, ensuring instruments move safely about the incision point \cite{kuo2009robotics}. This pivot point, typically aligned with the incision site, minimizes tissue trauma by preventing lateral forces that can cause bruising or tearing. Since its introduction, RCM-based technology has been widely adopted across various specialized surgical platforms, including the PRECEYES Surgical System (Preceyes BV, Eindhoven, The Netherlands) for vitreoretinal procedures, Yomi (Neocis Inc., Miami, FL, USA) for dental applications, and the da Vinci Surgical System (Intuitive Surgical Inc., Sunnyvale, CA, USA) for laparoscopic surgery.

Despite recent technological advances, RCM implementation continues to face practical challenges in surgical settings. Surgeons position trocars through incisions manually, relying on visual markers on the cannula to align the RCM with the body wall \cite{pandey2019robotics}, which can result in suboptimal placement. The RCM position relative to patient anatomy also shifts during procedures due to patient repositioning, robotic arm adjustments, or respiratory movement \cite{riviere2006robotic, rosa2015estimation, nasiri2024admittance}, causing unintended tissue forces at insertion points. While Intuitive Surgical's 2016 introduction of Integrated Table Motion for the da Vinci Xi system allows for patient position adjustments without undocking, these adjustments remain manually controlled by the surgical team, who may not detect misaligned RCM. The Senhance surgical system (Asensus Surgical, Durham, NC, USA) offers automatic pivot point identification at the beginning of a procedure, but only alerts surgeons to excessive forces at the incision during surgery \cite{nathan2021senhance} without indicating which direction the misalignment occurs.


\begin{figure}[t!]
    \centering
    \includegraphics[width=1\linewidth]{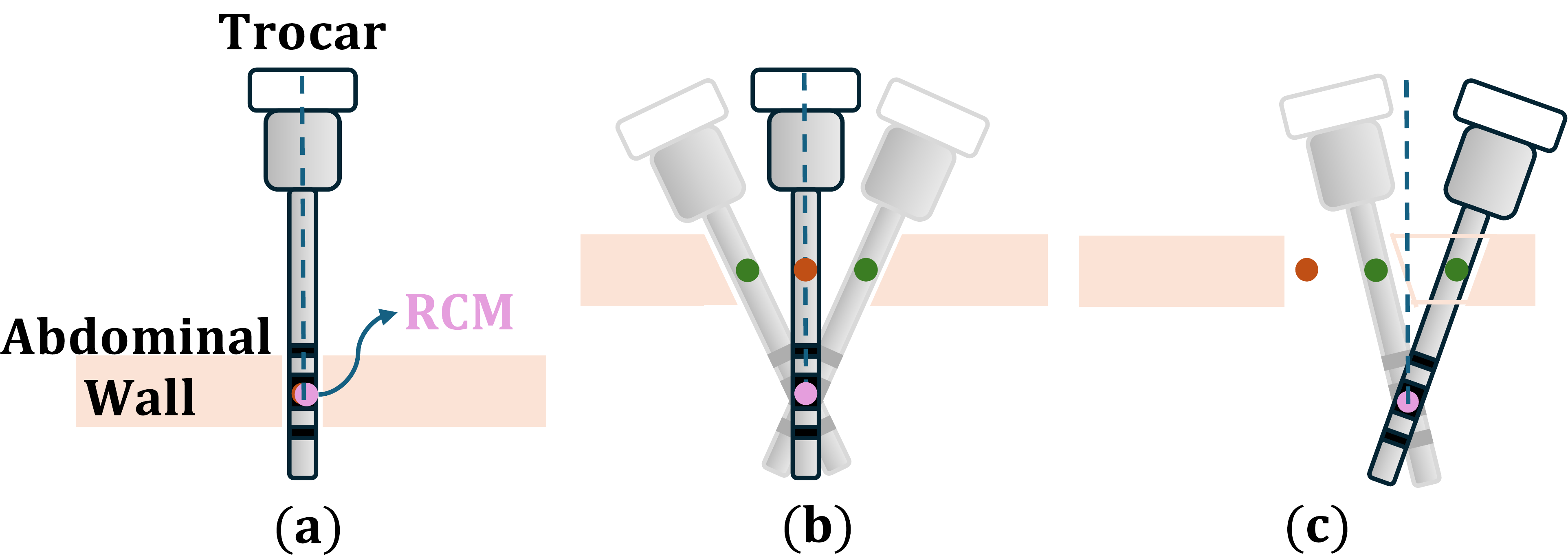}
    \vspace{-5mm}
    \caption{(a) With perfect trocar docking and exact alignment between the incision and the RCM (brown dot), tissue stretching is negligible. (b) Under pure coaxial misalignment, the incision is stretched as the robot pivots about the displaced RCM (pink dot). (c) When both coaxial and radial misalignments are present, the radial component introduces a constant offset that further increases tissue strain.}
    \label{fig:problem}
\end{figure}

This work addresses these gaps in robotic surgical safety by presenting a novel force estimation-based approach to dynamically estimate RCM misalignment during surgical procedures. In particular, our main contributions are: a) experimental quantification of forces exerted on soft tissue due to RCM misalignment, providing reference data for assessing clinical significance of positioning errors in robotic surgical systems; b) development of a machine learning–based framework for estimating forces acting on the surgical tool shaft without requiring additional sensors; and c) development of an RCM misalignment estimation method that leverages the predicted tool shaft forces, experimentally validated on the da Vinci Research Kit (dVRK) Classic~\cite{kazanzides-chen-etal-icra-2014}.

\section{Related Works}
\label{sec:related}
\subsection{RCM Identification Approaches}
Various methods have been developed to define RCMs in surgical robotics, broadly categorized as hardware-based (mechanical design) and software-based approaches. Correspondingly, different strategies have been investigated for estimating RCM positioning, including vision-based, sensor-based, and model-based approaches.

Vision-based methods typically employ optical tracking systems to identify the RCM point. Wilson \textit{et al.} evaluated RCM positioning using computer vision techniques that require dual cameras with direct visual access to the tool shaft, a configuration challenging to implement clinically due to workspace constraints and potential occlusion by surgical instruments \cite{wilson2010evaluating}. Wang \textit{et al.} determined the transformation between dual-arm robotic RCMs using only the endoscopic camera without external data sources \cite{wang2017vision}. While innovative, this approach only establishes relative transformations between dual-arm RCMs rather than maintaining safe RCM positioning relative to patient anatomy. Though theoretically adaptable for RCM repositioning, most vision-based methods function optimally in static environments and face practical limitations from occlusions and dynamic environmental changes.

Sensor-based approaches primarily leverage force/torque sensing to prevent unsafe tissue forces. Fontanelli \textit{et al.} developed a specialized force sensor for trocar tip attachment, but this cannot measure forces at the RCM~\cite{fontanelli2017novel}. Nasiri \textit{et al.} estimated and minimizes forces at the RCM using force/torque feedback at the robot base to adjust instrument positioning through an admittance controller \cite{nasiri2024admittance}. However, this approach was validated only on systems with software-based RCMs and requires additional sensor integration. 

Model-based approaches incorporate kinematic models and control algorithms to estimate RCM constraints. Dynamic model-based RCM estimation often requires external sensor measurements, as demonstrated in Nasiri \textit{et al.}'s extended work \cite{nasiri2024teleoperation} and Kastrisi \textit{et al.}'s admittance controller implementations \cite{kastritsi2021controller}. However, these approaches present integration challenges with mechanically-defined RCM systems due to limited built-in adaptability, complicating model-based correction implementation. Effective solutions will likely require hybrid approaches where software compensation addresses minor misalignments while significant deviations necessitate manual repositioning.

Despite their innovations, existing approaches inadequately address dynamic surgical environment challenges, including sudden force variations, while requiring additional sensors that complicate clinical deployment. These gaps motivate the development of an advanced force estimation-based RCM identification methodology, capable of real-time adaptation within existing surgical robotic platform constraints.

\subsection{Force Estimation Techniques}
Researchers have developed sensorless force estimation methods, broadly categorized into model-based and learning-based approaches. These sensorless methods primarily estimate free-space joint torque, from which forces can be calculated using the system's Jacobian matrix. Chua \textit{et al.} developed a vision and kinematics-based approach to estimate tip forces on the dVRK Classic, though their reliance on external cameras limited clinical applicability \cite{Chua2021CharacterizationOR}. Yilmaz \textit{et al.} proposed a multi-layer perceptron neural network to estimate free-space joint torque based on kinematics \cite{Yilmaz2020NeuralNB}. Wu \textit{et al.} subsequently extended this approach by implementing a Long Short-Term Memory (LSTM) neural network for improved estimation accuracy and introduced a secondary network to compensate for patient-robot interaction forces \cite{Wu2021RobotFE}. Building on this research trajectory, a hybrid model- and learning-based framework for dVRK Classic force estimation that combines the generalizability of model-based approaches with the adaptability of learning-based methods was developed~\cite{yang2024hybrid}. To the best of the authors' knowledge, none of these force estimation techniques have been specifically applied to determine RCM positioning.

\section{Methods}
\label{sec:method}
This section formalizes the geometric and mathematical representation of the RCM misalignment problem, and introduces a three-step framework for estimating both radial ($\delta_{r,x}$ and $\delta_{r,y}$) and coaxial ($D$) misalignments.

\subsection{Problem Formulation}
\begin{figure}[t!]
    \centering
    \vspace{+3mm}
    \includegraphics[width=1\columnwidth]{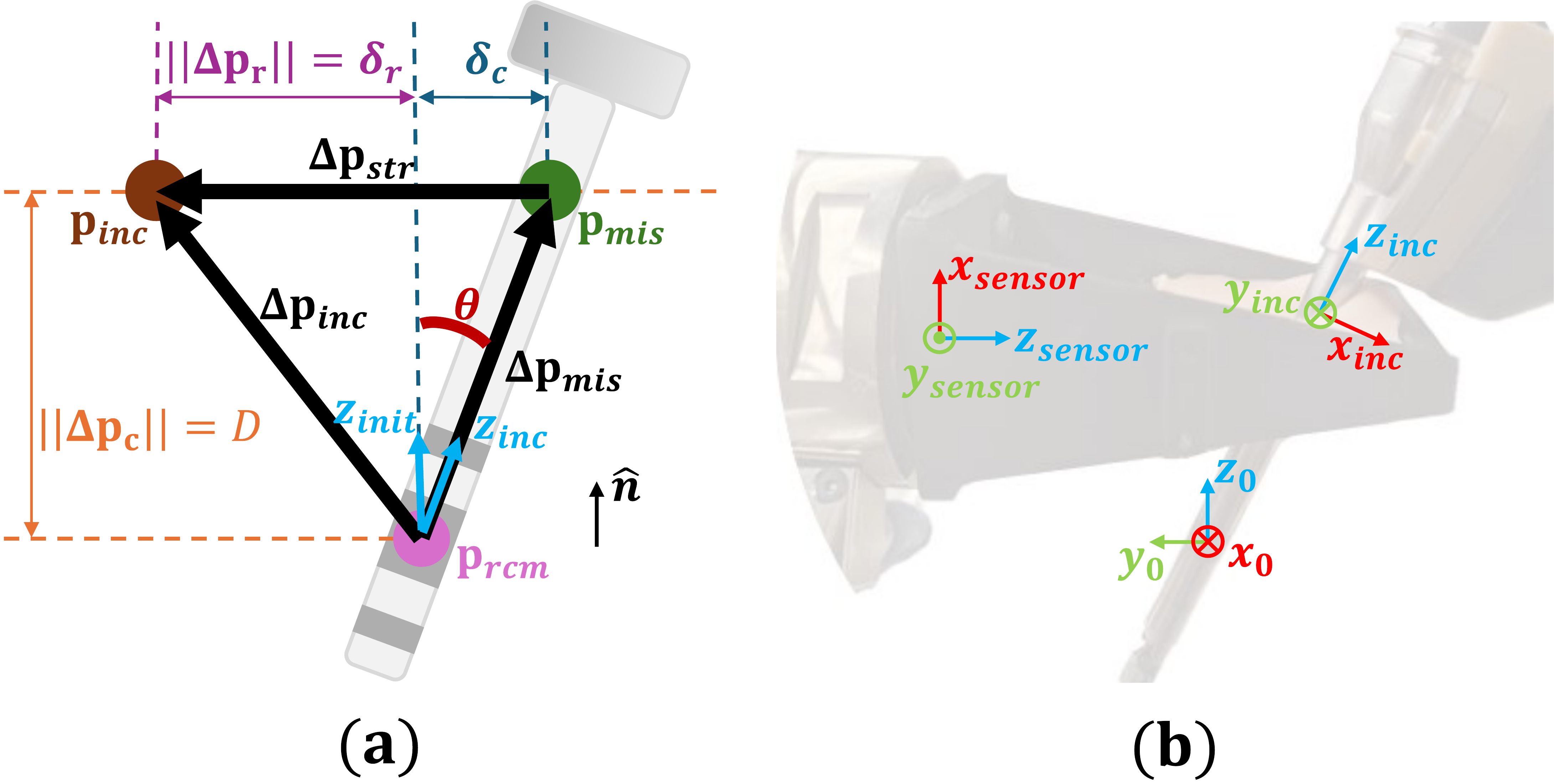}
    \caption{(a) Key Notations, Variables, and Parameters for RCM Misalignment Problem Formulation. (b) Illustration of the RCM misalignment corresponding frames assignment.}
    \label{fig:setup}
    
\end{figure}

The RCM misalignment problem is mathematically illustrated in Fig.~\ref{fig:setup}\,(a). Three points are of interest: the pink dot represents the robot RCM, located at the center of the thick black bar on the trocar, denoted $\mb{p}_{rcm} \in \realfield{3}$. The brown dot is the true incision point created at the start of surgery, denoted $\mb{p}_{inc} \in \realfield{3}$, and the green dot is the misaligned incision point after trocar displacement, denoted $\mb{p}_{mis} \in \realfield{3}$. 

Let the Euclidean displacement between every pair of points defined as $\Delta \mb{p}_{inc}$ = ($\mb{p}_{inc}- \mb{p}_{rcm}$), $\Delta \mb{p}_{mis}$ = ($\mb{p}_{mis} - \mb{p}_{rcm}$), and  $\Delta \mb{p}_{str}$ = ($\mb{p}_{inc} - \mb{p}_{mis}$). 
When the trocar is docked perpendicular to the patient's body surface and the RCM coincides with the incision point, the displacements satisfy
$\Delta\mb{p}_{rcm}$ = $\Delta \mb{p}_{inc}$ = $\Delta \mb{p}_{str}$ = 0. Under this condition, the incision experiences negligible stretching, and the interaction force on the abdominal wall is minimized, as shown in Fig.~\ref{fig:problem}\,(a). 

The initial trocar axis, \textit{i.e.} the $z$-axis of the RCM frame, is denoted by the unit vector $\uvec{z}_{init}$, normal to the local tissue surface. Let $\uvec{n}$ denote the surface normal unit vector of the local tissue at $\mb{p}_{rcm}$, with $\uvec{z}_{init} \parallel \uvec{n}$. We decouple the RCM misalignment into two modes:
\begin{itemize}[leftmargin=*]
    \item \textbf{Radial}: along the tangent plane of the local tissue surface, where $\Delta \mb{p}_r \perp \uvec{n}$ (see Fig.~\ref{fig:setup}\,(a)).
    \item \textbf{Coaxial}: along the normal direction of the local tissue surface, where $\Delta \mb{p}_c \parallel \uvec{n}$ (see Fig.~\ref{fig:setup}\,(a)).
\end{itemize}
The combined misalignment is thus expressed as $\Delta \mb{p}_{inc} = \Delta \mb{p}_r + \Delta \mb{p}_c.$

Radial and coaxial misalignments induce distinct components of tissue surface stretch. We denote the total stretch as $\Delta \mb{p}_{str}$ = $\delta_r \uvec{u}_r + \delta_c \uvec{u}_c$, where $\delta_r, \delta_c \in \realfield{}$ are the magnitude of radial and coaxial misalignments, respectively, and $\uvec{u}_r, \uvec{u}_c$ are unit vectors aligned with the directions of the resulting interaction forces.

Assuming that the unit vector $\uvec{n}$ remains unchanged throughout the surgery, we model the stretch caused by radial misalignment as $\delta_r = \|\Delta \mb{p}_r\|$. Since $\Delta \mb{p}_r \perp \uvec{n}$, we have:
\begin{equation}
\label{eq:delta_r_define}
    \Delta \mb{p}_r = \delta_r\uvec{u}_r = [{\delta}_{r,x}, {\delta}_{r,y}, 0],
\end{equation}
where ${\delta}_{r,x}, {\delta}_{r,y}$ are the components of the radial misalignment along the $x$- and $y$-axes of the initial incision frame. In addition, we model the stretch caused by coaxial misalignment as
\begin{equation}
\label{eq:delta_c_define}
    \delta_c = \|\Delta\mb{p}_{mis} - \left( \Delta\mb{p}_{mis} \cdot \uvec{n} \right)\uvec{n}\| = \|\Delta\mb{p}_{c}\| \tan\theta,
\end{equation}
where $\theta\in\realfield{}$ is the angle between the initial trocar axis $\uvec{z}_{init}$ and the trocar axis $\uvec{z}_{inc}$ after motion, as shown in Fig.~\ref{fig:setup}\,(a). Note that the resulting tissue stretch from the coaxial component is the projection of $\Delta\mb{p}_{mis}$ onto the local tissue surface plane. For simplicity, let ${D} = \|\Delta \mb{p}_c\|$. Since $\Delta \mb{p}_c \parallel \uvec{n}$, we have:
\begin{equation}
\label{eq:D_define}
    \Delta \mb{p}_c  = [0,0,D].
\end{equation}
Thus our vector of interest $\Delta \mb{p}_{inc}$ is composed of:
\begin{equation}
\label{eq:p_inc}
    \Delta \mb{p}_{inc} = [{\delta}_{r,x}, {\delta}_{r,y}, {D}].
\end{equation}

The external interaction force at the misaligned incision is modeled as:
\begin{equation}
\label{eq:cont_misalign_opt}
    \mb{f}_e = \mb{f}_{r} + \mb{f}_{c},
\end{equation}
where ${\mb{f}_r}=g(\delta_r)\uvec{u}_r$ and ${\mb{f}_c} = g(\delta_c)\uvec{u}_c$ represent the radial and coaxial components of this force, respectively. The function ${g(\cdot)}$ represents the mapping from tissue stretch to external force (${g(\cdot)}:\|\Delta\mb{p}_{str}\| \mapsto \mb{f}_e$). The force ${\mb{f}_e}$ can be directly related to the robot’s joint torques through the manipulator Jacobian:
\begin{equation}
\label{eq:overall_force}
     \mb{f}_{e} = \mb{J}_D\invT( \bs{\tau} - \bs{\tau}_{fs}).
\end{equation}
The term ${\mb{J}_D\invT}$ corresponds to the inverse transpose of the Jacobian matrix evaluated at the misaligned incision port.  $\bs{\tau}$ denotes the measured joint torque, while $\bs{\tau}_{fs}$ represents the free-space torque caused by gravity, friction, and other intrinsic factors. By combining Eq.~\ref{eq:cont_misalign_opt} and Eq.~\ref{eq:overall_force}, the following relationship is identified:
\begin{equation}
\label{eq:force_final_formula}
     g(\delta_r)\uvec{u}_r + g(\delta_c)\uvec{u}_c = \mb{J}_D\invT( \bs{\tau} - \bs{\tau}_{fs}).
\end{equation}

\begin{table}[t!]
    \begin{center}
    \vspace{+3mm}
    \caption{DH table of the Misaligned Incision Port, modified from~\cite{dvrkUserManual}.}
    \label{tab:dh_table}
    \normalsize
    \begin{tabular}{l l c c c c}
        \hline
        i  & Ref & $a_{i-1}$ & $\alpha_{i-1}$ & $d_i$ & $\theta_i$ \\
        \hline
        $1$ & $0$ & 0 & $+\frac{\pi}{2}$ & 0 & $q_1 + \frac{\pi}{2}$ \\
        $2$ & $1$ & 0 & $-\frac{\pi}{2}$ & 0 & $q_2 - \frac{\pi}{2}$ \\
        $inc$ & $2$ & 0 & $-\frac{\pi}{2}$ & $-D\sec\theta$ & 0 \\
        $sensor$ & $0$ & 0 & $\frac{\pi}{2}$ & 0 & $\frac{\pi}{2}$ \\
        \hline
    \end{tabular}
    \end{center}
    \footnotesize {Note: Ref stands for the reference link frame of link i. $a_{i-1}$, $\alpha_{i-1}$, $d_i$ and $\theta_i$ are the modified DH parameters of link $i$. $q_i$ is the joint position of link $i$. Frames ‘inc’ and ‘sensor’ denote the incision port frame and the force sensor frame, respectively.}
    \vspace{-2mm}
\end{table}

To compute the Jacobian at the incision port, we first adjust the kinematics by modifying the Denavit-Hartenberg (DH) table. The robot base frame is defined at the RCM, as described in the dVRK user manual~\cite{dvrkUserManual}. Our analysis focuses on the DH parameters of the first three joints, as only these joints are affected by RCM misalignment. The resulting incision port frame is illustrated in Fig.~\ref{fig:setup}\,(b), and the corresponding parameters are listed in Table~\ref{tab:dh_table}. The Jacobian at the incision is thus given by:
\begin{align} 
\label{eq:jacobian}
\mb{J}_D &=
\resizebox{0.9\linewidth}{!}{$
\begin{bmatrix}
\Rot{0}{1} \uvec{u}_{z_0} \times (\mb{d}_0^{inc} - \mb{d}_0^1) &
\Rot{0}{2} \uvec{u}_{z_0} \times (\mb{d}_0^{inc} - \mb{d}_0^2) &
\Rot{0}{inc} \uvec{u}_{z_0}
\end{bmatrix}
$} \notag \\[6pt]
&=
\begin{bmatrix}
- D \sec\theta \, c_1 c_2 & D \sec\theta \, s_1 s_2 & c_2 s_1 \\
0 & D \sec\theta \, c_2 & -s_2 \\
- D \sec\theta \, c_2 s_1 & - D \sec\theta \, c_1 s_2 & -c_1 c_2
\end{bmatrix},
\end{align}
where \(\uvec{u}_{z_0} = \begin{bmatrix} 0, 0, 1 \end{bmatrix}^T\),  
$\Rot{0}{i}$ is the rotation matrix and $\mb{d}_0^i$ is the position vector of frame $i$ with respect to the robot base frame.

We can also express $\theta$ as a function of joint values $\mb{q}$. Let the initial joint positions be $\mb{q}_0 = [q_{1_0}, q_{2_0}, q_{3_0}]^\mathrm{T}$, and the joint positions at time $t$ be $\mb{q}_t = [q_{1_t}, q_{2_t}, q_{3_t}]^\mathrm{T}$. We can represent $\uvec{z}_{init}$ and $\uvec{z}_{inc}$ as:
\begin{align}
\uvec{z}_{init} = 
\begin{bmatrix}
\cos q_{2_0} \sin q_{1_0} \\
-\sin q_{2_0} \\
-\cos q_{1_0} \cos q_{2_0}
\end{bmatrix}, \,
\uvec{z}_{inc} = 
\begin{bmatrix}
\cos q_{2_t} \sin q_{1_t} \\
-\sin q_{2_t} \\
-\cos q_{1_t} \cos q_{2_t}
\end{bmatrix}. \notag
\end{align}
Therefore, $\theta$ is derived as:
\begin{align}
    \theta &= \arccos (\uvec{z}_{init}\uvec{z}_{inc}) \nonumber \\
    &= \arccos ( \cos q_{2_0} \, \cos q_{2_t}\cos(q_{1_t}-q_{1_0}) \nonumber \\
    &\qquad \qquad  + \sin q_{2_0} \, \sin q_{2_t}).
\end{align}
In Eq.~\ref{eq:force_final_formula}, the only unknown quantities on the right-hand side are now the coaxial misalignment $D$ and free space torque $\bs{\tau}_{fs}$. The Jacobian term $\mb{J}_D$ can be expressed as a function of $D$ and the joints configuration $\mb{q}$, while the total joints torques $\bs{\tau}$ can be directly measured. The following section describes the procedure for estimating $\bs{\tau}_{fs}$.

\subsection{Dynamic Free-space Torque Identification}

The relationship between $\bs{\tau}_{fs}$ and robot motion can be identified via the robot’s free-space inverse dynamics model:
\begin{equation}
    \bs{\tau}_{fs} = \bs{f}(\mb{q}, \mb{\dot q}),
\end{equation}
where $\mb{q}, \mb{\dot q}$ are the joint positions and velocities and $\bs{f}(\cdot)$ maps robot motion to joint torque, representing the inverse dynamics. Previous works have shown that such a mapping could be parameterized by either model-based or learning-based methods~\cite{Wu2021RobotFE, Wang2019ACO}. In this work, we adopt the learning-based method of Wu \textit{et al.}~\cite{Wu2021RobotFE}, chosen for the generalizability of its LSTM model and stability of its performance. We train the LSTM network to estimate free-space torques for each joint of the dVRK Classic Patient Side Manipulator (PSM) using joint angles and velocities as inputs. A quadratic loss function is used to compare the estimated and measured free-space torques. 

\subsection{Three-Step RCM Misalignment Estimation}
\label{Sec:scheme}
We present in this subsection the optimization workflow for estimating $\Delta\mb{p}_{inc}$ in Eq.~\ref{eq:p_inc}. Since there are three unknowns but only a single equation relating them (Eq.~\ref{eq:force_final_formula}), we design a three-step optimization workflow to decouple the estimation of misalignment components. By controlling the experimental setup, each step isolates specific conditions that make it possible to estimate the unknown variables independently, as described below.


\textbf{1. Baseline Data Collection:} We assume that at the beginning of the surgery, the trocar is docked perfectly perpendicular to the tissue surface and the RCM coincides with the incision point, as shown in Fig.~\ref{fig:problem}\,(a). The robot joint positions $\mb{q}_0 = [q_{1_0}, q_{2_0}, q_{3_0}]^\mathrm{T}$ are recorded for subsequent use as a reference configuration in radial misalignment estimation.

\textbf{2. Tissue Model Identification:} We intentionally introduce a controlled coaxial offset of $\sim$25-35\,mm to characterize the relationship between trocar misalignment and the forces acting on the misaligned incision port, as shown in Fig.~\ref{fig:problem}\,(b). This step is clinically viable, as surgeons using the da Vinci system often insert the trocar deeper to access specific anatomies. In our experiments, the ground-truth coaxial misalignment values are measured directly, whereas in actual surgery known misalignments can be introduced using the setup joint.

Under controlled pure coaxial misalignment where $D^*$ is known and $[\delta^*_{r,x}, \delta^*_{r,x}]$ = [0, 0], we collect joint data $\{ \mb{q}_c, \dot{\mb{q}}_c, \bs{\tau}_c \}$ while moving the robot for 1–2 minutes. Let $\bs{\tilde \tau}_{c, {fs}}$ denote the estimated free-space torque in this condition. Using the recorded robot data, we estimate $\bs{\tilde \tau}_{{c},{fs}} = \bs{f}(\mb{q}_c, \mb{\dot q}_c)$, compute both $D^*\tan\theta$ and $\mb{J}_{D^*}\invT( \bs{\tau}_c - \bs{\tilde \tau}_{c, fs})$, and then identify the mapping function:
\begin{equation}
\label{eq:g_func}
    {g(\cdot)}:D^*\tan\theta \mapsto \mb{J}_{D^*}\invT( \bs{\tau}_c - \bs{\tilde \tau}_{c, fs})
\end{equation}
via symbolic regression. Specifically, we supply numerical pairs of $D^*\tan\theta$ and $\mb{J}_{D^*}\invT( \bs{\tau}_c - \bs{\tilde \tau}_{c, fs})$, and search over candidate function families (\textit{e.g.}, linear, polynomial, sinusoidal, exponential). We perform this search with a genetic algorithm–based optimization method, Symbolic Regression in Python (PySR)~\cite{Cranmer2023InterpretableML}.

\textbf{3. Hybrid Misalignment Estimation:}
After identifying $g(\cdot)$, we now address the estimation of the ``hybrid" misalignment vector $\Delta\mb{p}_{inc}$, composed of both coaxial and radial components. This step integrates dynamic and static measurements to first separate the radial and coaxial force contributions, and then solve for the corresponding misalignment magnitudes and directions.

\paragraph{Formulation of Total External Force From Dynamic Measurements}
With both radial and coaxial misalignments present and unknown, we collect joint data $(\mb{q}_h, \mb{\dot q}_h, \bs{\tau}_h)$ during 1–2 minutes of robot motion, as shown in Fig.~\ref{fig:problem}\,(c). Using the recorded data, the overall estimated external force $\mb{\tilde f}_e$ can be expressed as a function of the unknown coaxial displacement $D$:
\begin{equation}
\label{eq:hybrid_est}
    \mb{\tilde f}_e  = 
    \mathbf{J}_D\invT ( \bs{\tau}_h - \bs{\tilde \tau}_{h, {fs}}),
\end{equation}
with $\bs{\tilde \tau}_{h, {fs}} = \bs{f}(\mb{q}_h, \mb{\dot q}_h)$ being the free-space torque.

\paragraph{Decoupling Force Components From Static Measurements}
To separate the contributions of coaxial and radial misalignments, the robot is first returned to the baseline configuration $\mb{q}_0$ established in Step 1. Although $\mb{q}_0$ was originally recorded when the RCM coincided with the incision point, patient motion during surgery may shift the tissue relative to the robot, so offsets can still be present even when the robot returns to the same configuration. The trocar is assumed to be perpendicular to the patient’s surface, so only the effect of radial misalignment remains. The corresponding joint torques $\bs{\tau}_{r_0}$ are then recorded, and the resulting interaction force arises solely from the radial offset:
\begin{equation}
\label{eq:static_est}
    \mb{\tilde f}_r  = 
    \mathbf{J}_D\invT ( \bs{\tau}_{r_0} - \bs{\tilde \tau}_{{r_0}, {fs}}) = \mathbf{J}_D\invT \Delta\bs{\tilde \tau}_r,
\end{equation}
where $\bs{\tilde \tau}_{{r_0}, {fs}} = \bs{f}(\mb{q}_0, 0)$, and $\Delta\bs{\tilde \tau}_r$ represents the difference in joint torque between the initial docking and misaligned condition. Subtracting this radial component from the total force in Eq.~\ref{eq:cont_misalign_opt} decouples the estimated coaxial component:
\begin{equation}
\label{eq:coaxial_est}
    \mb{\tilde f}_c  = 
    \mathbf{J}_D\invT ( \bs{\tau}_{h} - \bs{\tilde \tau}_{{h}, {fs}}) - \mathbf{J}_D\invT \Delta\bs{\tilde \tau}_r,
\end{equation}
enabling independent estimation of coaxial and radial forces.

\paragraph{Estimation of Coaxial Misalignment via Numerical Optimization}
The coaxial force component satisfies ${\mb{\tilde f}_c} = g(\delta_c)\uvec{u}_c$. Combining it with Eq.~\ref{eq:delta_c_define}, we obtain:
\begin{equation}
    g(D\tan\theta)\uvec{u}_c  = 
    \mathbf{J}_D\invT ( \bs{\tau}_{h} - \bs{\tilde \tau}_{{h}, {fs}} -\Delta\bs{\tilde \tau}_r).
\end{equation}
We finally formulate the following constrained optimization problem to solve for $\tilde D$:
\begin{equation}\label{eq:3}
\begin{split}
    \tilde D = \argminB_{D}& \quad  \tfrac{1}{2} \psi^\mathrm{T}\psi, \\
    \text{s.t.} & \quad -20\,\text{mm} \leq D \leq 70\,\text{mm}
\end{split}  
\end{equation}
where:
\begin{equation}\label{eq:loss_func}
\psi = g(D\tan \theta) -  \| \mathbf{J}_D\invT ( \bs{\tau}_h - \bs{\tau}_{h, {fs}} - \Delta\bs{\tau}_r) \| .
\end{equation}

\paragraph{Radial Misalignment Recovery}
With $\tilde D$ determined, the coaxial component of misalignment is given by $\Delta\mb{\tilde p}_{inc} = [0,0,\tilde D]$. The radial component can then be derived from Eq.~\ref{eq:static_est}, since ${\mb{\tilde f}_r}=g(\delta_r)\uvec{u}_r$. Substituting the optimized value of $\tilde D$ yields:
\begin{equation}
\label{eq:radial}
    g(\delta_r)\uvec{u}_r  = \mathbf{J}_{\tilde D}\invT \Delta\bs{\tilde \tau}_r.
\end{equation}
Referring to Eq.~\ref{eq:delta_r_define}, the radial misalignment vector is calculated as:
\begin{align}
\label{eq:delta_r}
\Delta\mb{\tilde p}_r = g^{-1} \big(\|\mathbf{J}_{\tilde D}\invT \Delta\bs{\tilde \tau}_r\|\big) \,\frac{\mathbf{J}_{\tilde D}\invT \Delta\bs{\tilde \tau}_r}{\|\mathbf{J}_{\tilde D}\invT \Delta\bs{\tilde \tau}_r\|}.
\end{align}
The overall hybrid misalignment is obtained by combining coaxial and radial components $\Delta\mb{\tilde p}_{inc} = \Delta\mb{\tilde p}_c + \Delta\mb{\tilde p}_r$, and the complete procedure is summarized in Algorithm~\ref{alg:optimization}.

It is important to note that this methodology relies on three key assumptions:
\begin{itemize}[leftmargin=*]
\item The presence of misalignment is detected, but $\Delta\mb{p}_{inc}$ remains unknown.
\item The tissue surface normal $\uvec{n}$ does not change appreciably before and after the misalignment, \textit{i.e.}, the displaced surface is nearly parallel to the original.
\item A nonzero coaxial misalignment is always present and sufficiently large to induce measurable joint torque changes.
\end{itemize}


\begin{algorithm}[htbp]
\caption{Three-Step RCM Misalignment Estimation}
\label{alg:optimization}
\begin{algorithmic}[1]
\State \textbf{Input:} $\bs{\tau}, \mb{q}, \dot{\mb{q}}, D_0$
\State \textbf{Output:} $\Delta \mb{\tilde p}_{inc} = [\tilde{\delta}_{r,x}, \tilde{\delta}_{r,y}, \tilde{D}]$

\State \textbf{Step 1: Baseline} Dock trocar RCM perpendicular at incision, record $\mb{q}_0$ and free-space torque $\bs{\tilde \tau}_{r_0, fs}$

\State \textbf{Step 2: Coaxial Identification} Introduce known $D^*$, set $\delta_r^*=0$, move robot and collect $\{\mb{q}_c, \dot{\mb{q}}_c, \bs{\tau}_c\}$  
\State Estimate $\bs{\tilde \tau}_{c, fs} = f(\mb{q}_c, \dot{\mb{q}}_c)$, compute $\mb{J}_{D^*}^{-T} (\bs{\tau}_c - \bs{\tilde \tau}_{c, fs})$  
\State Fit mapping ${g(\cdot)}:D^*\tan\theta \mapsto \mb{J}_{D^*}\invT( \bs{\tau}_c - \bs{\tilde \tau}_{c, fs})$ via symbolic regression (PySR)

\State \textbf{Step 3: Hybrid Misalignment Estimation}
\State Move robot under unknown misalignment, collect $\{\mb{q}_h, \dot{\mb{q}}_h, \bs{\tau}_h\}$  
\State Estimate $\bs{\tilde \tau}_{h, fs} = f(\mb{q}_h, \dot{\mb{q}}_h)$, compute total force as a function of $D$, $\mb{f}_e = \mathbf{J}_D^{-T} (\bs{\tau}_h - \bs{\tilde \tau}_{h, fs})$  

\State Return robot to baseline configuration $\mb{q}_0$, measure $\bs{\tau}_{r_0}$  
\State Estimate $\bs{\tilde \tau}_{r_0, fs} = f(\mb{q}_0, 0)$, compute radial force as a function of $D$, $\mb{f}_r = \mathbf{J}_D^{-T} (\bs{\tau}_{r_0} - \bs{\tilde \tau}_{r_0, fs})$  

\State Compute coaxial force as a function of $D$, $\mb{f}_c = \mb{f}_e - \mb{f}_r$  

\State Take $D_0$ as initial guess, use nonlinear optimizer to solve $\tilde D = \arg\min_D \frac{1}{2} \| g(D \tan \theta) - \|\mb{f}_c\| \|^2$  

\State Recover coaxial misalignment: $\Delta \mb{\tilde p}_c$ = [0,0,$\tilde D$], radial misalignment: $\Delta \mb{\tilde p}_r = g^{-1}(\|\mb{f}_r\|) \frac{\mb{f}_r}{\|\mb{f}_r\|}$     

\State \textbf{Return:} $\Delta \mb{\tilde p}_{inc} = \Delta \mb{\tilde p}_c + \Delta \mb{\tilde p}_r$ 
\end{algorithmic}
\end{algorithm}

\section{Experimental Setup}
\label{sec:experimental_setup}

We build a 3-DoF aluminum frame with an adjustable moving platform, as shown in Fig. \ref{fig:setup}. The platform integrates an ATI force sensor (ATI Industrial Automation, Apex, NC, USA) for measuring ground-truth forces and a 3D-printed tissue holder that secures a 1\,cm thick, 9$\times$9\,cm$^2$ silicone pad. The pad is fabricated from EcoFlex-50 (Smooth-On Inc., Macungie, PA, USA) to mimic body wall interactions at the trocar insertion site. As shown in prior studies~\cite{Chaves2022AbdominalWT, Gallarello2019PatientSpecificAP, Lavazza2023StrainRT}, the selected material and thickness represent reasonable approximations for simulating abdominal wall interactions in phantom-based experiments. Using this setup, we obtain ground-truth measurements of both incision misalignment ($D^*$) and interaction forces at the incision ($\mb{f}^*_e$) to validate our optimization results.


\subsection{Forces Induced by Trocar Misalignment}
\label{sec:demo_nece_exp_setup}

We begin by measuring the range of forces that can result from trocar misalignment. The trocar RCM is docked at the tissue incision port, after which a coaxial misalignment is introduced. The robot is pivoted to a specified angle $\theta$, held statically, and the corresponding ground-truth interaction force is recorded for each combination of $D$ and $\theta$ to construct a force–angle relationship. The tested conditions include $D\,\text{=}\,(0\,\text{mm}, 10\,\text{mm}, 20\,\text{mm}, 30\,\text{mm}, 40\,\text{mm})$ and $\theta\,\text{=}\,(-45^{\circ}, -30^{\circ}, -15^{\circ}, 0^{\circ}, 15^{\circ}, 30^{\circ}, 45^{\circ})$. As we expect the radial force to be independent of robot motion, we consider only coaxial misalignment in this experiment. 

\subsection{Free Space Torque Estimation}

The free-space training dataset for the dVRK PSM is collected through teleoperation for $\sim$15 minutes, during which the robot is moved without touching the environment. The dataset was split into 70\% for training, 20\% for validation, and 10\% for testing. The neural network architecture, hyperparameter settings, and hardware devices follow the configurations reported in previous studies~\cite{hao2024effectiveness, yang2024hybrid}.

\subsection{Interaction Force Estimation at Misaligned Incision Port}
\label{sec:misalign_setup}
This experiment evaluates whether interaction forces at the incision port can be accurately estimated, and whether the effect of radial and coaxial misalignments can be effectively decoupled within those forces. To address these questions, the following three experiments are proposed.

\subsubsection{Force Estimation under Pure Coaxial Misalignment}
\label{subsec:pure}
We move the robot to its home position $\mb{q}_0\,\text{=}\,[0^{\circ},0^{\circ},120\,\text{mm}]$ and dock the trocar RCM at the phantom incision port. We then introduce a pure coaxial misalignment $\Delta \mb{p}_{inc_{(1)}}^* \,\text{=}\, [0\,\text{mm},\,0\,\text{mm},\,30\,\text{mm}]$ and teleoperate the robot for 1 minute while recording joint positions, velocities, and torques. We compute force estimates using Eq.~\ref{eq:overall_force} and \ref{eq:jacobian}, and compare them with ATI sensor data using root mean squared error (RMSE).

\subsubsection{Force Estimation under Hybrid Misalignment}
\label{subsec:hybrid}
We apply a hybrid offset $\Delta \mb{p}_{inc_{(2)}}^* \,\text{=}\, [10\,\text{mm},\,15\,\text{mm},\,30\,\text{mm}]$ and reproduce the trajectory from Experiment (a) while recording joint data. We estimate forces using Eq.~\ref{eq:hybrid_est} and \ref{eq:jacobian}, and compare them with ATI sensor data.

\subsubsection{Decoupling of Coaxial Force Contribution}
Under $\Delta \mb{p}_{inc_{(2)}}^*$, we move the robot back to $\mb{q}_0$. In this static condition, we record joint torques and compute the radial force contribution using Eq.~\ref{eq:static_est} and ~\ref{eq:jacobian}. We then subtract this constant radial force from the forces in Sec.~\ref{subsec:hybrid} to obtain the coaxial component, referring to Eq.~\ref{eq:coaxial_est}. Next, we calculate the RMSE between estimated pure coaxial force from~\ref{subsec:pure} and decoupled coaxial force from~\ref{subsec:hybrid}. We also calculate the RMSE between the measured coaxial components from the ATI force sensor. We expect the pure coaxial force from~\ref{subsec:pure} to match the decoupled coaxial force from~\ref{subsec:hybrid}, yielding a small RMSE, since~\ref{subsec:pure} and~\ref{subsec:hybrid} share the same coaxial misalignment and robot trajectory. Thus, a small RMSE supports our hypothesis that the radial force acts as a constant offset.

\subsection{RCM Misalignment Estimation}
\label{sec:optimize_setup}

The following pipeline is adopted according to the proposed three-step scheme to validate our method and assess its ability to estimate and decouple trocar misalignments. 

\subsubsection{Initial Docking}
The trocar is first docked perpendicular to the surface of the abdominal wall phantom, with its black reference bar, representing the robot’s RCM, aligned with a predefined incision port at the phantom’s center. The robot joint positions are recorded as $\mb{q}_0\,\text{=}\,[0^{\circ},0^{\circ},120\,\text{mm}]$.

\subsubsection{Tissue Model Identification}
A pure coaxial misalignment of $D=30$\,mm is introduced, and a 1-minute dataset recorded during tele-operation. This dataset is then used with PySR to identify the function $g(\cdot)$ shown in Eq.~\ref{eq:g_func}. The candidate function library is chosen as $\{ 1,\, x,\, x^2,\, x^3,\, \sin(x),\, \cos(x),\, \exp(x) \}$ including constant, polynomial, and nonlinear terms to capture the behavior of the phantom tissue. The function with the highest score, determined by the best trade-off between loss reduction and model complexity, is selected~\cite{Cranmer2023InterpretableML}.

\subsubsection{Hybrid Misalignment Estimation}
In this step, we validate Algorithm~\ref{alg:optimization}. We collect several new test dataset with $D\,\text{=}\,(10\,\text{mm}, 20\,\text{mm}, 30\,\text{mm}, 40\,\text{mm}, 50\,\text{mm})$ and $[\delta_{r,x}, \delta_{r,y}]\,\text{=}\,([0\,\text{mm}, 0\,\text{mm}], [10\,\text{mm}, 15\,\text{mm}])$. The robot is teleoperated under these conditions, and joint data recorded for evaluation. The optimization procedure is implemented in MATLAB using \texttt{lsqnonlin}~\cite{DennisNonlinear} as the solver and the \texttt{Trust-Region-Reflective} algorithm~\cite{Coleman1994Reflective, Coleman1993AnIT} for optimization. Convergence is declared when the relative change in the sum of squared residuals falls below $10^{-6}$, which serves as the stopping criterion. Estimation accuracy is quantified by the error $e = \|\Delta \mathbf{p}_{inc} - {\Delta \mathbf{\tilde p}}_{inc}\|$.

\section{Results}
\label{sec:experiment}

\subsection{Forces Induced by Trocar Misalignment}
\begin{figure}[b!]
    \centering
\includegraphics[width=0.85\linewidth]{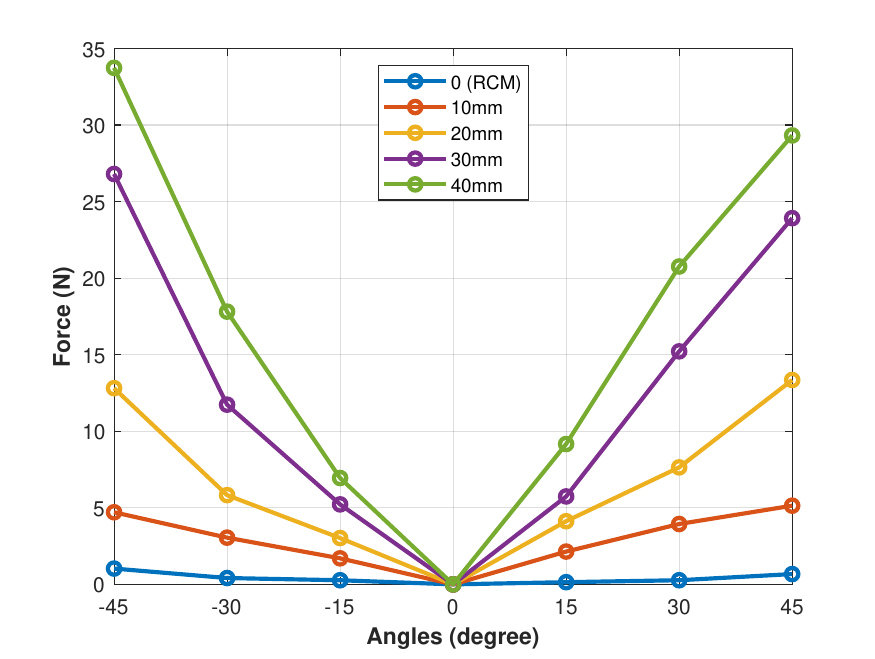}
    \caption{Static force-angle plot in different coaxial misalignments and pivoting angles.}
    \label{fig:neccesity}
\end{figure}

When the incision point coincides with the RCM, the interaction forces are expected to be zero or negligible. With increasing coaxial misalignment, larger values of $\theta$ produce higher interaction forces, as $\delta^*_c=D^*\tan\theta$ increases accordingly. The results are shown in Fig.~\ref{fig:neccesity}. 

The experiments demonstrate that when the incision port is displaced more than 20\,mm from the RCM, the resulting interaction force exceeds 12\,N. For $(D^*,\theta)$\,=\,(40\,mm, 45$^{\circ}$), the force reaches 35\,N, a magnitude that poses a risk of abdominal tissue damage~\cite{Kriener2023MechanicalCO, Miller2023QuantitativeTO, Konerding2011MaximumFA, Frstemann2011ForcesAD}. These findings underscore the importance of detecting and correcting RCM misalignment in da Vinci robot–assisted surgery.

\subsection{Free-space Torque Estimation}
Force estimation accuracy in free space was validated following the procedure outlined in~\cite{Yilmaz2020NeuralNB}. As summarized in Table~\ref{tab:intrinsic}, the model achieves normalized root mean square error (NRMSE) below 6\,\% in estimating free-space torque, comparable to the results reported in~\cite{Yilmaz2020NeuralNB}.

\begin{table}[t!]
\centering
\vspace{+3mm}
\footnotesize
\caption{NRMSE (\%) of intrinsic dynamics estimation for proposed method (PM) compared to~\cite{Yilmaz2020NeuralNB} for the full six PSM joints.}
\begin{tabular}{|c|c|c|c|c|c|c|}
\hline
Method & $\tau_1$ & $\tau_2$ & $f_3$ & $\tau_4$ & $\tau_5$ & $\tau_6$ \\ 
\hline\hline
~\cite{Yilmaz2020NeuralNB} & 4.4 & 3.5 & 5.0 & 4.7 & 6.3 & 6.8 \\
PM & 1.6 & 1.5 & 5.3 & 4.5 & 4.8 & 5.5 \\
\hline
\end{tabular}
\vspace{-2mm}
\label{tab:intrinsic}
\end{table}

\subsection{Interaction Force Estimation at Misaligned Incision Port}
\label{sec:exp_int_fe}
\begin{figure}[b!]
    \centering
    \includegraphics[width=1\linewidth]{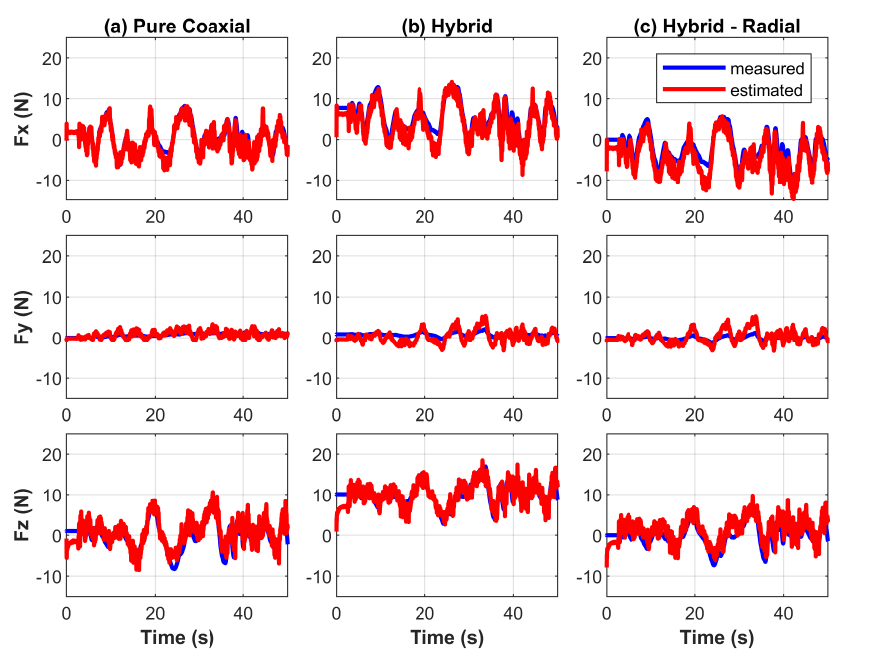}
    \caption{(a) Force estimation at the incision port with $\Delta \mb{p}_{inc_{(1)}}$. (b) Force estimation at the incision port with $\Delta \mb{p}_{inc_{(2)}}$. (c) Decoupled coaxial force from (b) after subtracting the constant offset corresponding to the radial misalignment force.}
    \label{fig:incision_est_combine}
\end{figure}

Fig.~\ref{fig:incision_est_combine} shows the estimated interaction forces at the incision port under different misalignment conditions, and Table~\ref{tab:incision_est} shows the quantitative force estimation accuracy. These errors are comparable to or lower than previously reported tool-tip force estimation methods~\cite{Wu2021RobotFE, hao2024effectiveness, yang2024hybrid}.

To analyze the decoupling of misalignments, we extracted the coaxial component of the hybrid experiment by subtracting the radial contribution derived from Eq.~\ref{eq:radial}. The resulting decoupled coaxial forces (Figure~\ref{fig:incision_est_combine}c) closely matched the pure coaxial condition (Figure~\ref{fig:incision_est_combine}a).  The quantitative RMSE between pure coaxial and decoupled coaxial conditions is [2.8\,N, 1.0\,N, 1.2\,N] for the ATI sensor, and [4.4\,N, 1.5\,N, 1.8\,N] for the estimator in $Fx$, $Fy$, and $Fz$, respectively.
 
\begin{table}[htbp]
\centering
\vspace{3mm}
\caption{RMSE of interaction force, ground truth vs. estimated. $\Delta \mb{p}_{inc_{(1)}}^*$=[0\,mm, 0\,mm, 30\,mm], $\Delta \mb{p}_{inc_{(2)}}^*$=[10\,mm, 15\,mm, 30\,mm]}
\label{tab:incision_est}
\begin{tabular}{ccccc}
\hline
\multicolumn{1}{|c|}{\textbf{Method}} & \multicolumn{1}{c|}{\textbf{Location}} & \multicolumn{1}{c|}{$Fx(N)$} & \multicolumn{1}{c|}{$Fy(N)$} & \multicolumn{1}{c|}{$Fz(N)$} \\ \hline\hline
\multicolumn{1}{|c|}{LSTM} & \multicolumn{1}{c|}{Incision, $\Delta \mb{p}_{inc_{(1)}}$} & \multicolumn{1}{c|}{1.4} & \multicolumn{1}{c|}{0.7} & \multicolumn{1}{c|}{2.1} \\ \hline
\multicolumn{1}{|c|}{LSTM} & \multicolumn{1}{c|}{Incision, $\Delta \mb{p}_{inc_{(2)}}$} & \multicolumn{1}{c|}{2.0} & \multicolumn{1}{c|}{1.4} & \multicolumn{1}{c|}{1.8} \\ \hline
\multicolumn{1}{|c|}{~\cite{Wu2021RobotFE}*} & \multicolumn{1}{c|}{Tool Tip} & \multicolumn{1}{c|}{2.3} & \multicolumn{1}{c|}{1.5} & \multicolumn{1}{c|}{3.3} \\ \hline
\multicolumn{1}{|c|} {~\cite{hao2024effectiveness}*} & \multicolumn{1}{c|}{Tool Tip} & \multicolumn{1}{c|}{1.1} & \multicolumn{1}{c|}{0.9} & \multicolumn{1}{c|}{0.9} \\ \hline
\end{tabular}
\end{table}


\subsection{RCM Misalignment Estimation}

Following the setup in Section~\ref{sec:optimize_setup}, the identified highest-scoring functions is:
\begin{equation}
\label{eq:g_mapping}
    y = 788\,x,
\end{equation}
suggesting that $g(\cdot)$ is approximately linear, with interaction forces nearly proportional to the misalignment. Under the input uncertainty bound of $D\pm 3$\,mm, the optimal models consistently yield a linear relationship, with multipliers ranging from 651 to 972. In the next section, we use Eq.~\ref{eq:g_mapping} for misalignment optimization.



Table~\ref{tab:optimizeD_delta} summarizes the recovered misalignments. For pure coaxial misalignments above 20\,mm, error between estimates and ground truth are below 6\,mm. Hybrid misalignment errors also remain below 6\,mm when $D > 20$\,mm. The worst hybrid case is $[10,15,10]$\,mm, where the coaxial term is overestimated, leading to $e=17.2$\,mm.

Overall, the results show that the proposed optimization framework provides reliable estimation once misalignment exceeds 20\,mm, with good decoupling of radial and coaxial components. 


\begin{table}[htbp]
\centering\caption{Optimization results for misalignment vector $\Delta \mathbf{p}_{inc} = [\delta_{r,x}, \delta_{r,y}, D]$.}
\label{tab:optimizeD_delta}
{
\begin{tabular}{|c|c|c|}
\hline
$\Delta \mathbf{p}_{inc}^*$\,(mm) & ${\Delta \mathbf{\hat p}}_{inc}$\,(mm) & $e$\,(mm) \\ \hline
$[0, 0, 10]$  & $[4.4, -2.4, 25.0]$   & 15.8  \\ \hline
$[10, 15, 10]$  & $[5.3, 12.4, 26.3]$   & 17.2  \\ \hline
$[0, 0, 20]$  & $[4.3, -3.5, 29.3]$ & 10.8  \\ \hline
$[10, 15, 20]$& $[8.4, 8.8, 34.0]$   & 15.2 \\ \hline
$[0, 0, 30]$  & $[0.4, -0.3, 30.2]$  & 0.5  \\ \hline
$[10, 15, 30]$& $[9.3, 9.7, 34.4]$   & 5.6  \\ \hline
$[0, 0, 40]$  & $[-1.7, -1.0, 37.0]$ & 3.7  \\ \hline
$[10, 15, 40]$& $[9.2, 11.0, 44.7]$  & 5.7  \\ \hline
$[0, 0, 50]$  & $[0.1, 1.9, 44.3]$ & 6.0 \\ \hline
$[10, 15, 50]$& $[9.2, 12.8, 52.9]$  & 3.7  \\ \hline
\end{tabular}%
}
\end{table}

\subsection{Discussion}
\label{sec:discussion}
The results highlight three key findings. First, even relatively small RCM misalignment (20\,mm) can induce trocar–patient interaction forces exceeding 10\,N at large angles, underscoring the importance of an accurate misalignment estimation framework. Second, we demonstrate that both radial and coaxial force components can be estimated independently without direct sensing, using a learning-based approach that achieves errors below 5\,N. Third, the proposed three-step optimization scheme reliably estimates RCM misalignment under coaxial displacements greater than 20\,mm, with errors below 5 mm. Notably, this performance generalizes beyond the training condition: although the tissue modeling function $g(\cdot)$ is learned at $D=30$\,mm, estimation results remain accurate at $D=40, 50$\,mm.

A key limitation arises within 20\,mm of the RCM, where force estimation errors increase because joint torque measurements are less sensitive to the small interaction forces in this range. As a result, learning-based force estimation is less effective.
Consistent with our experience, sensorless approaches, whether model-based or learning-based, struggle when external forces are below approximately 10\,N. This limitation is particularly relevant in the present context, as Fig.~\ref{fig:neccesity} demonstrates that trocar interaction forces within 20\,mm of the RCM remain in this low-force regime. More importantly, prior studies have indicated that forces below 10\,N are unlikely to cause patient injury~\cite{Kriener2023MechanicalCO, Miller2023QuantitativeTO, Konerding2011MaximumFA, Frstemann2011ForcesAD}, suggesting that the range in which estimation errors increase corresponds to a clinically acceptable safety margin. This observation is further explained by the kinematics: when $\theta=0$ or $D=0$, interaction forces are nearly zero due to negligible tangential displacement, and they increase progressively with larger $\theta$ and $D$. Importantly, within 20\,mm of the RCM, interaction forces stay below 12\,N. Interestingly, the da Vinci trocar exhibits a similar tolerance zone, marked by a thick black bar at the RCM and two thinner bars approximately 17\,mm above and below. Although their precise function is undocumented, prior work suggests they serve as guides for trocar–incision alignment~\cite{Qafiti2023NotDE, Higuchi2011Robotic}. Our results align with this interpretation, indicating that coaxial misalignments within this range do not generate clinically significant forces.

We acknowledge that the proposed framework is subject to important limitations, which we link here to the three assumptions outlined in Section~\ref{Sec:scheme}, Step 4. First, the framework assumes that misalignment is detected, but that its magnitude and direction are unknown. In principle, such detection could be achieved intraoperatively through vision-based or torque-based polling methods; however, these mechanisms are beyond the present scope. Second, we assume that the tissue surface remains approximately parallel before and after motion. This assumption underpins the derivation of Eq.~\ref{eq:delta_c_define}, Eq.~\ref{eq:delta_r_define}, and Eq.~\ref{eq:p_inc}. In practice, patient motion may not be purely translational; rotational components can alter the surface normal vector $\uvec{n}$, invalidating the relation $\delta_c= D \tan \theta$. Third, we assume that coaxial misalignment is always present and sufficiently large to induce measurable changes in joint torques. As observed, the framework is effective beyond the clinically safe region; however, within $\pm 20$\,mm of the RCM, joint torques become increasingly insensitive to external forces. This decay in sensitivity is expected: as the incision port approaches the RCM, the lever arm shortens, reducing motor torques. Concurrently, the third column of the Jacobian at the incision tends toward zero, potentially reducing its rank. At the RCM itself, the Jacobian becomes singular and non-invertible. Consequently, with the current configuration (dVRK fixed without a setup joint), the framework cannot resolve pure radial misalignments using joint torques alone. Future work will focus on addressing these challenges, by potentially incorporating additional observables, such as setup joint torque measurements to aid in identifying $\delta_r$.



\section{Conclusion}
We propose a sensorless framework to optimize for the RCM misalignment. Through our experiments, we show that misalignment can potentially generate enough force to damage tissue. For misalignments larger than 20\,mm, our proposed algorithm is effective at estimating the misalignment.  Accurately estimating the RCM misalignment is a step toward automation compensation and, thus, reducing damage to tissue at the incision point.


\bibliographystyle{IEEEtran}
\bibliography{references}

\end{document}